\documentclass[10pt, conference, compsocconf]{IEEEtran}

\usepackage{graphics} 
\usepackage{times} 
\usepackage{amsmath} 
\usepackage{amssymb}  

\usepackage[ruled, linesnumbered]{algorithm2e}
\usepackage{caption}
\usepackage{subcaption}
\usepackage[acronym]{glossaries}
\usepackage{sidecap}
\usepackage{wrapfig}
\usepackage{tabularx, booktabs, ragged2e}
\usepackage{adjustbox}
\usepackage{multirow}
\usepackage{makecell}
\usepackage{gensymb}
\usepackage{comment}
\usepackage{xr}
\usepackage{lipsum}  
\usepackage{arydshln}
\usepackage{mathrsfs}
\usepackage[pagebackref=true,breaklinks=true,letterpaper=true,colorlinks,bookmarks=false, citecolor=blue]{hyperref}

\newcommand{\ie}{\textit{i}.\textit{e}., }
\newcommand{\eg}{\textit{e}.\textit{g}., }
\newcommand{\etal}{\textit{et} \textit{al}. }

\newacronym{gan}{GAN}{generative adversarial network}
\newacronym{ffsgan}{FFS-GAN}{Frontal Face Synthesizing GAN}
\newacronym{tpgan}{TP-GAN}{Two-pathway GAN}
\newacronym{capggan}{CAPGGAN}{Couple-Agent Pose-Guided GAN}

\newacronym{d}{$D$}{discriminator}
\newacronym{g}{$G$}{generator}

\begin{document}
%
\title{Dual-Attention GAN for Large-Pose Face Frontalization}


\author{\IEEEauthorblockN{Yu Yin, Songyao Jiang, Joseph P. Robinson, Yun Fu}
\IEEEauthorblockA{Northeastern University, Boston, MA\\
\{yin.yu1, jiang.so, robinson.jo\}@northeastern.edu, yunfu@ece.neu.edu}
}

%


\maketitle

\begin{abstract}
Face frontalization provides an effective and efficient way for face data augmentation and further improves the face recognition performance in extreme pose scenario. Despite recent advances in deep learning-based face synthesis approaches, this problem is still challenging due to significant pose and illumination discrepancy. 
In this paper, we present a novel Dual-Attention Generative Adversarial Network (DA-GAN) for photo-realistic face frontalization by capturing both contextual dependencies and local consistency during GAN training. 
Specifically, a self-attention-based generator is introduced to integrate local features with their long-range dependencies yielding better feature representations, and hence generate faces that preserves identities better, especially for larger pose angles.
Moreover, a novel face-attention-based discriminator is applied to emphasize local features of face regions, and hence reinforce the realism of synthetic frontal faces. Guided by semantic segmentation, four independent discriminators are used to distinguish between different aspects of a face (\ie skin, keypoints, hairline, and frontalized face).
By introducing these two complementary attention mechanisms in generator and discriminator separately, we can learn a richer feature representation and generate identity preserving inference of frontal views with much finer details (i.e., more accurate facial appearance and textures) comparing to the state-of-the-art.
Quantitative and qualitative experimental results demonstrate the effectiveness and efficiency of our DA-GAN approach.
\end{abstract}


\IEEEpeerreviewmaketitle

\section{Introduction}
Automatic face understanding from imagery is, and has been, a popular topic throughout the research community. Modern-day, data-driven models have pushed state-of-the-art on increasingly challenging benchmark datasets~\cite{Cao18, kemelmacher2016megaface, whitelam2017iarpa, yin2019joint}, with face-based models deployed in markets that span social-media, attribute understanding~\cite{wang2017kinship}, and more. A challenge that persists, however, is that of extreme poses-- face-based models tend to breakdown on samples of faces that are viewed at extreme angles, pitches, and yaws. The task of face frontalization corrects for this by aligning faces captured at a side-view to the front. Thus, face frontalization is a task that serves to enhance facial recognition as a preprocessing step. Additionally, this task could serve as a means of data augmentation. Furthermore, the same models could be used to align faces for practical purposes (\eg photo albums or commercial products).

Typical face frontalization methods~\cite{hu2018pose,huang2017beyond, li2019m2fpa} use only on conv-layers. Since nodes in a conv-layer are connected only to a small local neighborhood of nodes in the prior layer, it's difficult and inefficient to compute long-range dependencies using conv-layers alone.
Considering the large pose discrepancy between two views of a face, we introduce a self-attention modules in \gls{g} that capture long-range contextual information yielding better feature representations, and hence generate more faces that best preserves identities, and especially for larger poses. 

\begin{figure}[t]
    \centering
    \includegraphics[trim=0in 0.6in 0in 0in,clip,width=0.9\linewidth]{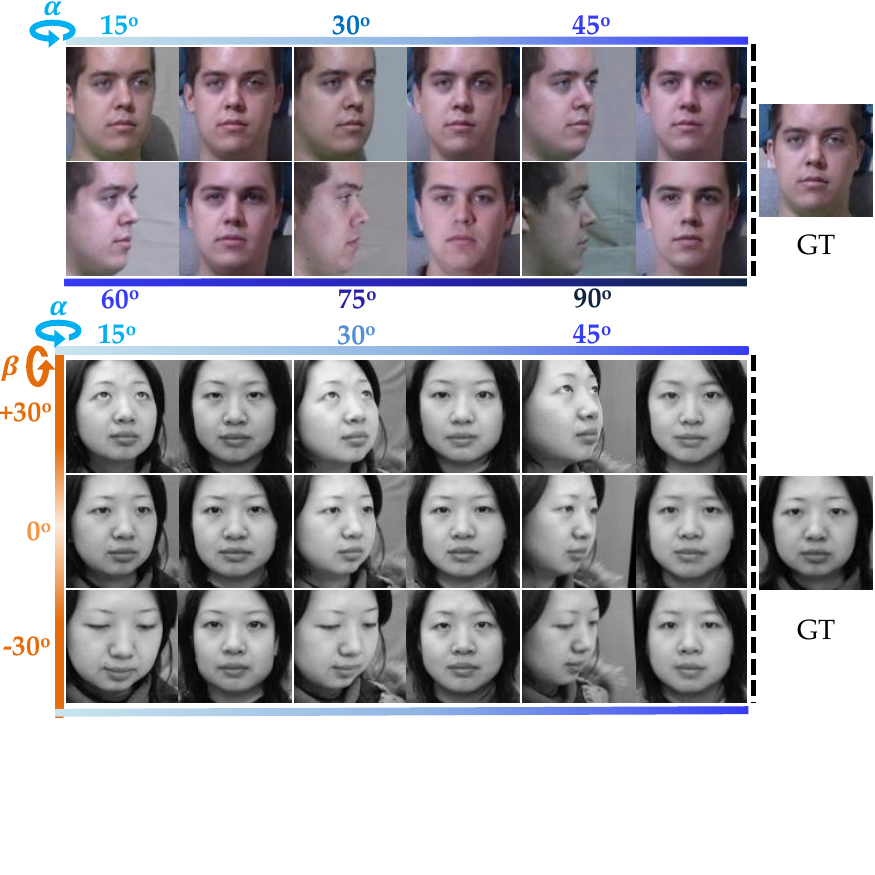}
    \caption{\textbf{Synthesized results of DA-GAN.} 
    Top two rows show the input side-view face images and our frontalized results for Multi-PIE; bottom three rows show the input yawed and pitched faces with our frontalized faces for CAS\_PEAL\_R1. Their ground-truth frontal faces are shown on the right side. }
    \label{fig:samples}
    \vspace{-3mm}
\end{figure}

Existing methods tend to distinguish on a generated image as a whole, but are tolerant on its finer details, which leads to unexpected artifacts on the synthesized results. Small artifacts might be acceptable for other applications. But for face applications, people are extremely sensitive to any small distortions that they may feel pretty unsettling to artificial faces due to the Uncanny Valley Effect \cite{mori2012uncanny}. To synthesize photo-realistic frontal faces, it requires the generator to pay attention to finer details and avoid generating artifact. 
So we propose an additional mechanism called face-attention to \gls{d}, which yields improved photo-realism with added discriminators that focus on particular regions of the face (\ie along with the \gls{d} for entire face, three additional discriminators are trained using pre-defined, masked out regions of the face. We dub the proposed model as Dual-Attention Generative Adversarial Network (DA-GAN). \footnote{The code is available at: https://github.com/YuYin1/DA-GAN.}

The benefits of DA-GAN are as follows. First, the added attention mechanisms in both \gls{g} and \gls{d} work in a complimentary fashion. Specifically, the self-attention in \gls{g}, added to the top-most and second-topmost layers, enables the model to capture long-term dependencies in image space, providing a means to preserve the true identity of the subject-- this is essential when deployed as pre-processing for facial recognition, which we demonstrate the effectiveness experimental using renowned face recognition benchmark data. To the best of our knowledge, we are the first to apply self-attention in \gls{g} for this problem. Furthermore, the face-attention in \gls{d} is a novel scheme that uses additional discriminators to provide more gradients (\ie signal) to learn by at training. Thus, face-attention faces four discriminators off against \gls{g} via adversarial training, and with each discriminator attending to a different aspect of the face (Fig.~\ref{fig:framework}). Ablation studies show that each \gls{d} compliments one another, providing overall improved performance with frontalized faces of higher quality (Section~\ref{sec:facesynthesis} and ~\ref{sec:ablation}). As we demonstrate, the different \gls{d} making up face-attention improves the particular facial regions for which it focuses (\eg $D_{h}$ focusing on the hairline and, thus, provides improved synthesized imagery in the respective region). Furthermore, identity is preserved with the addition of a facial recognition network trained to recognize subject identity (Section~\ref{sec:identity}). 

We make three key contributions in this work.
\begin{enumerate}
    \item A self-attention \gls{g} is introduced to capture long-range contextual dependencies, yielding better feature representations for preserving true identity of the subject.
    \item A face-attention \gls{d} is employed to enforce local consistency and improve synthesized imagery in particular facial regions. We further show that each component in \gls{d} compliments one another, providing overall improved performance with frontalized faces of higher quality.
    \item We show both quantitative and qualitative results to demonstrate that the proposed DA-GAN significantly outperforms the state-of-the-art methods, especially under extreme poses (\eg $90^\circ$).
\end{enumerate}

\section{Related work}
\subsection{GAN}
One of the machinery that takes the research community by storm is \gls{gan}~\cite{goodfellow2014generative}, which uses an adversarial learning scheme to leverage \gls{d} against \gls{g} such that both sides improve over training. The training of GANs is analogous to a two-player game between G and D, which has been widely used in image generation. 
Benefiting from recent advances in GAN models, notable achievement have been made for face frontalization.
\gls{tpgan}~\cite{huang2017beyond} is the first to propose a two-channel approach for frontal face synthesis, which is capable of capturing local details and comprehending global structures simultaneously. Shortly thereafter, \cite{shen2018faceid} develops a GAN-based framework that recombines different identities and attributes to preserve identities when synthesizing faces in an open domain.
After that, in \cite{zhao2018towards}, pose invariant feature extraction and frontal face synthesis are learned jointly in a way to benefit one another. 

\subsection{Face Frontalization}
Face frontalization is a computer vision task aiming to align faces at various views to a canonical position (\ie frontal). Progresses have been made through 2D/3D texture mapping~\cite{ferrari2016effective, hassner2015effective, zhu2015high, jeni2016person}, statistic modeling~\cite{sagonas2015robust, sagonas2015face, booth20163d, booth2018large} and deep learning-based methods~\cite{cao2018learning, huang2017beyond, yim2015rotating, yin2017towards, zhao2018towards, zhang2019pose}. For instance, Hassner \etal\cite{hassner2015effective} employs one single and unmodified 3D facial shape to reference all query images to frontalize faces. By solving a constrained low-rank minimization
problem, a statistical frontalization model is proposed to joint align and frontalize faces~\cite{sagonas2015robust}.

Recently, deep convolution neural networks (CNN) have proven it's powerful capability on face frontalization. A disentangled representation learning GAN (DR-GAN) is proposed in ~\cite{tran2017disentangled} to learn a generative representation, which is explicitly disentangled from other face variations (\eg pose).
FF-GAN~\cite{yin2017towards} is a \gls{gan} founded on a 3D facial shape model as a reference to handle cases of extreme posed faces in the wild. 
\cite{zhao2018towards} then propose PIM as an extension of TP-GAN. Specifically, the improvement is a strategy for domain adaption that improve recognition performance on faces with extreme pose variations.
The proposed DA-GAN differs from the existing works by incorporating attention mechanisms in both \gls{g} and \gls{d}. 
Two different types of attention mechanism are employed to compliment one another, providing overall improved performance with fronalized faces of higher quality and better preserved identity.

\begin{figure*}[t]
    \centering
    \includegraphics[trim=0in 0.6in 0in 0.05in,clip,width=0.85\linewidth]{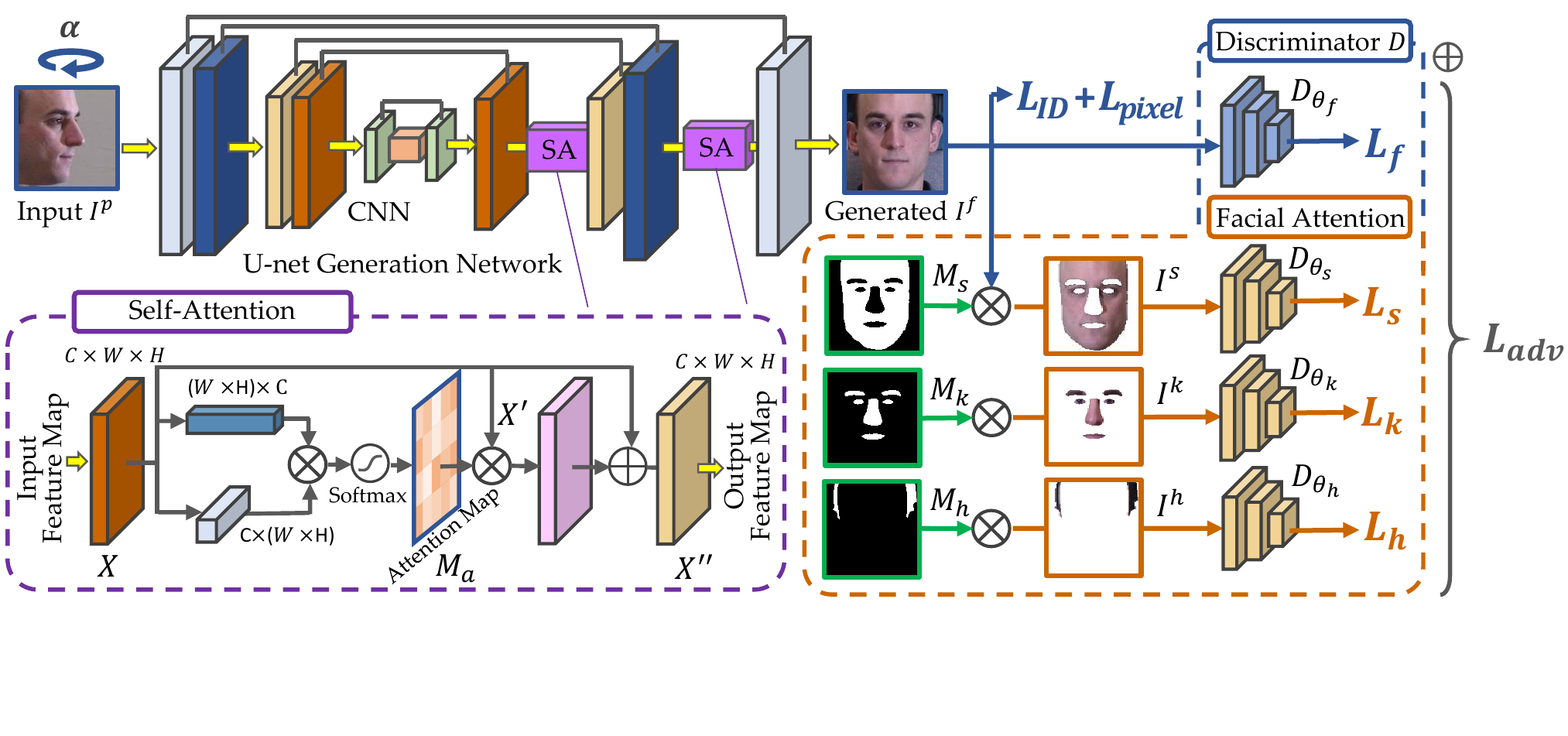}
    \caption{\textbf{Proposed framework.} DA-GAN consists of a self-attention \gls{g} and a face-attention \gls{d}. The self-attention in \gls{g} computes the response at a position as a weighted sum of the features in every spatial location to help capturing long-range contextual information. The face-attention in \gls{d} is based on four independent discriminator models (\ie $D_{f}$, $D_{s}$, $D_{k}$, $D_{h}$) to enforce local consistency between $I^p$ and $I^f$. Additionally, pixel similarity loss and identification loss are employed to help generate photo-realistic and identity preserving frontal faces.
    }
    \label{fig:framework}
    \vspace{-3mm}
\end{figure*}

\subsection{Attention and Self-attention}
The attention mechanism, broadly speaking, mimicks human sight by attempting to learn as we perceive: human perception avoids saturation from information overload by honing in on features that commonly relate to an entity of interest.
Attention are first used in recurrent neural nets for image classification~\cite{mnih2014recurrent}. 
Then in 2017, self-attention is introduced in \cite{vaswani2017attention} for machine translation tasks.
Generally speaking, self-attention, is an attention mechanism that captures dependencies at different positions of a single sequence without recurrent calculations. Recently, it has been shown to be very useful in computer vision tasks such as image classification~\cite{woo2018cbam, xie2018rethinking}, image generation~\cite{zhang2018self, mejjati2018unsupervised}, and scene segmentation~\cite{zhang2018self, zhao2018psanet}.
Different from existing work, our DA-GAN employ two different types of attention to jointly capture long-range dependencies and local features.

\section{Methodology}
In this section, we first give a definition of the face frontalization problem and define the symbols used in our methodology. Then we talk about the framework structure of proposed DA-GAN and how the dual attention mechanism contribute to the frontalization results. After that, we provide objective functions to optimize the networks. 

\subsection{Problem Formulation}
Let $\mathbb{P}_{data}$ be a dataset which contains frontal and side-view facial images. Let $\{I^f, I^p\}$ be a pair of frontal and side-view face images of a same person sampled from $\mathbb{P}_{data}$. Given a side-view face image $I^p$, our goal is to train a generator \gls{g} to synthesize the corresponding frontal face image $\hat{I}^f=G\left(I^p\right)$, which is expected to be identity preserving and visually faithful to $I^f$.

To achieve this, we propose DA-GAN shown in Fig.~\ref{fig:framework} to train the target generator \gls{g}. DA-GAN has two main components, the self-attention \gls{g} and face-attention discriminator (\gls{d}). The self-attention in \gls{g} captures long-range contextual information yielding better feature representations.
Meanwhile, face-attention in \gls{d} is based on four independent discriminator models, with each targeting different characteristics of a face. Hence, it helps to enforce local consistency of $I^p$ and $I^f$.
In this way, our model is able to generate frontal view images closer to the ground-truth and exhibit photo-realistic and identity preserving faces.

\subsection{Self-attention in \gls{g}}
\glsreset{g}
Inspired by U-Net~\cite{ronneberger2015u}, our \gls{g} consists of a encoder-decoder structure with skip connections for multi-scale feature fusion. A self-attention module is added to the last two feature maps of size $64\times64$ and $128\times128$, respectively. The detailed architecture of the \gls{g} is provided in the supplementary material.

Considering the illumination discrepancy between frontal and side-view face images resulted from large pose angles, we introduced a self-attention module in \gls{g} to capture the long-range contextual information for better feature representations. 
Typically, nodes in a convolutional layers are only computed from a small local neighborhood of nodes from the previous layer. It is difficult and inefficient when computing long-range dependencies with convolutional layers alone. With self-attention, the response at a position is computed as a weighted sum of all features from different spatial locations and, hence, it bridges long-range dependencies for any two positions of the feature maps, and information for the non-linear transformation.

Given a feature map $X\in\mathbb{R}^{C \times H \times W}$, we first generate an attention map $M_a\in\mathbb{R}^{N \times N}$ by calculating the inter-relationship of the feature map, where $N=H \times W$ (Fig~\ref{fig:framework}). For this, the feature is fed to two different $1\times1$ convolutional layers to generate two new feature maps $A, B\in\mathbb{R}^{C \times H \times W}$. Then, we reshape $A, B$ to $\mathbb{R}^{C \times N}$ and perform matrix multiplication to $A$ and $B^\top$, respectively, where the superscript $\top$ denotes matrix transpose. Finally, the weights are normalizes using softmax. The attention map is computed as
\begin{equation}\label{eq:gene_att_map}
M_a = \sigma(A^\top \cdot B),
\end{equation}
where $\sigma$ denotes the softmax function, and $f(\cdot)$ and $g(\cdot)$ denote the two different $1\times1$ convolutional layers.

Meanwhile, the original feature $X$ is fed to a convolutional layers and reshaped to $\mathbb{R}^{C \times N}$ to generate a new feature map $X'$. Then, $X'$ is multiplied by the attention map $M$ and reshaped to $\mathbb{R}^{C \times H \times W}$. 
Finally, we multiply $M$ by a scalar parameter, which is then added to the original feature $X$. The output $X''\in\mathbb{R}^{C \times H \times W}$ is calculated as
\begin{equation}\label{eq:add_att_map}
    X''_j = X_j + \mu \sum_{i=1}^N M_{ji}X'_i,
\end{equation}
where $i,j$ are positions of the maps, and $\mu$ is a scalar parameter initialized as 0 and adapted during training.

\subsection{Face-attention in \gls{d}}
To synthesize photo-realistic frontal faces, the generative models have to pay attention to every single detail beyond distinguishing on the whole face. So we further introduce a novel face-attention scheme by employing three additional segmentation-guided discriminators, which collaborate with the discriminator \gls{d}$_f$, but focus on different local regions of the faces. Specifically, we divide frontal faces into three local regions (skin, keypoints, and hairline), and assign each region to a regional discriminator ($D_s$, $D_k$, and $D_h$). Each regional discriminator tends to improve the synthesized imagery in respective region and compliments one another.

We parse frontal faces into three predefined regions inspired by~\cite{li2019m2fpa}. Specifically, we use a pre-trained model~\cite{liu2015multi} as an off-the-shelf face parser $f_P$ to generate three masks, and then apply them on the frontal face image to create regional images, which are a low-frequency region $I^s$ (\ie skin regions), key-point features $I^k$ (\ie eyes, brows, nose, and lips), and the hairline $I^h$. Mathematically speaking,
\begin{equation}\label{eq:frontal}
M_s, M_k, M_h = f_P(I^f),
\end{equation}
where $M_s, M_k, M_h$ are the masks of skin, key-point features and hairline regions. Their subscripts remain consistent with the signals. Thus,
\begin{align}\label{eq:masked}
&{\textit{real}}&I^s= I^f{\odot} M_s, ~I^k=I^f{\odot} M_k, ~I^h = I^f{\odot} M_h;\nonumber\\
&{\textit{fake}}&\hat{I}^s=\hat{I}^f{\odot} M_s, ~\hat{I}^k=\hat{I}^h{\odot} M_k, ~\hat{I}^h=\hat{I}^f{\odot} M_h \text{.}
\end{align}
where ${\odot}$ is the element-wise product, and $\hat{I}^s, \hat{I}^k, \hat{I}^h$ represent regional images of skin, keypoint and hairline.

Respectively, four discriminators ($D_f$, $D_s$, $D_k$ and $D_h$) try to distinguish between the real frontal face images of four views ($I^f$, $I^s$, $I^k$ and $I^h$) and their corresponding synthesized frontal face images ($\hat{I}^f$, $\hat{I}^s$, $\hat{I}^k$ and $\hat{I}^h$) following their superscripts. All these discriminators are trained with the generator $G$ adversarially. Thus, the proposed face-attention consists of four independent adversarial losses of four independent discriminators,
\begin{equation}\label{eq:minmix}
 \mathcal{L}_j = \mathbb{E}_{I^j} \Big [\log D_f(I^j) \Big ] +\mathbb{E}_{\hat{I}^j} \Big [\log(1 - D_j(\hat{I}^j)) \Big ],
\end{equation}
where $j\in\{f, s, k, h\}$. Each $D_j$ tries to maximize its objective $\mathcal{L}_j$ against $G$ that tries to minimize it. The full objective can be expressed using a min-max formulation:
\begin{eqnarray}
    \min_{G} \max_{D} \mathcal{L}_\mathit{adv}(D,G),
    \label{eq::gan-problem}
\end{eqnarray}
where $\mathcal{L}_{adv}$ is the overall adversarial loss that

\begin{eqnarray}  
\begin{aligned}
    \mathcal{L}_{adv}=&\sum_{j\in\{f, s, k, h\}}\mathcal{L}_j(D_j,G)\\
    =&\sum_{j\in\{f, s, k, h\}}\Big (\mathbb{E}_{I^j}\left[\log D_j(I^j)\right]\\
    &~~~~~~~~~~~~~~+\mathbb{E}_{\hat{I}^j}[\log(1 - D_j(\hat{I}^j))] \Big), 
\end{aligned}
\end{eqnarray}
where $j\in\{f, s, k, h\}$ produce losses $\mathcal{L}_f$, $\mathcal{L}_s$, $\mathcal{L}_k$, and $\mathcal{L}_h$, respectively. Each of them tends to improve synthesized imagery in respective region and compliments one other. 

\begin{figure}[t]
    \centering
    \includegraphics[trim=0in 0.2in 0in 0in,clip,width=0.85\linewidth]{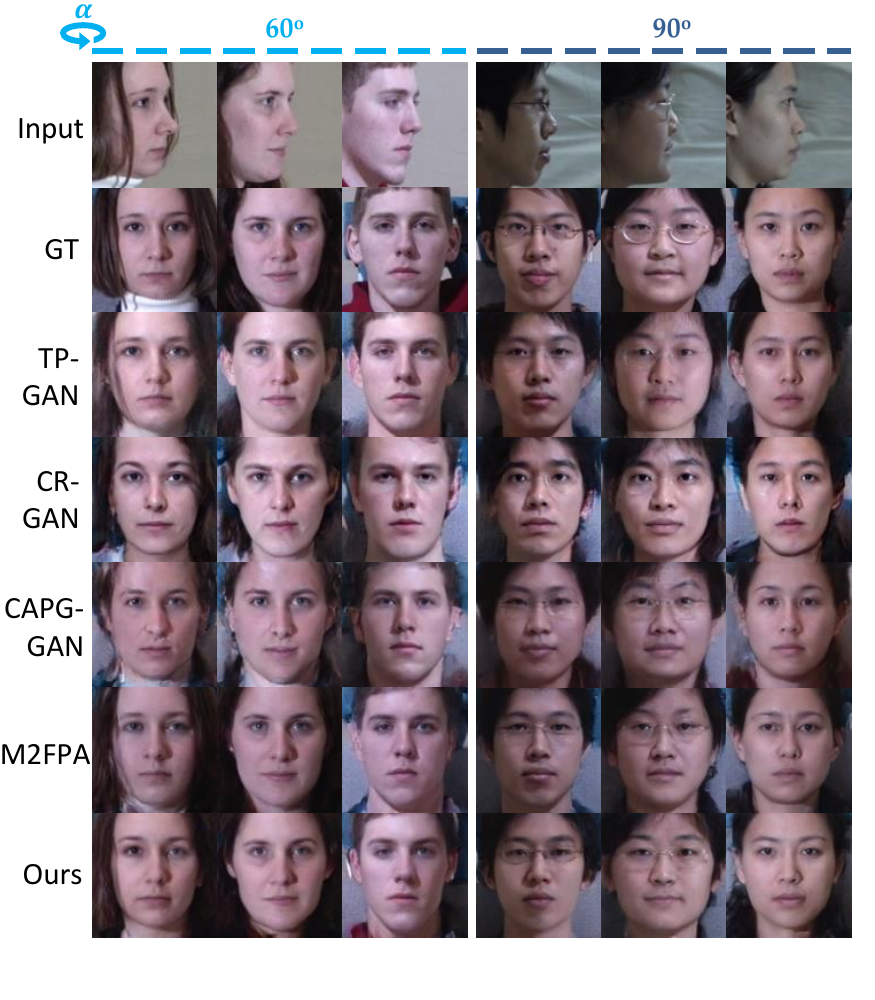}
    \caption{\textbf{Multi-PIE results.} Comparison with SOTA across extreme yaw ($\alpha$) poses. DA-GAN recovers frontal faces with finer details (\ie more accurate facial shapes and textures).}
    \label{fig:compare_state_of_the_art}
    \vspace{-8mm}
\end{figure}

\subsection{Objective Function of \gls{g}}

\subsubsection{Identity Preserving Loss}
 A critical aspect of evaluating face frontalization is the preservation of identities during the synthesis of frontal faces. We exploit the ability of pre-trained face recognition networks to extract meaningful feature representations to improve the identity preserving ability of \gls{g}. Specifically, we employ a pre-trained 29-layer Light CNN\footnote{Downloaded from \href{https://github.com/AlfredXiangWu/LightCNN}{https://github.com/AlfredXiangWu/LightCNN}.}~\cite{wu2018light} with its weights fixed during training to calculate an identity preserving loss for \gls{g}. The identity preserving loss is defined as the feature-level difference in the last two fully connected layers of Light CNN between the synthesized frontal face and the ground-truth frontal face:
 \begin{equation}
    \mathcal{L}_{ID} = \sum_{i=1}^2 ||p_i(I^f)-p_i(\hat{I}^f)||^2_2
 \end{equation}
where $p_i(\cdot) (i\in{1,2})$ are the output features from the fully connected layers of Light CNN, and $||\cdot||_2$ is the L2-norm.

\begin{figure}[!t]
    \includegraphics[trim=0in 0.04in 0in 0in,clip,width=.9\linewidth]{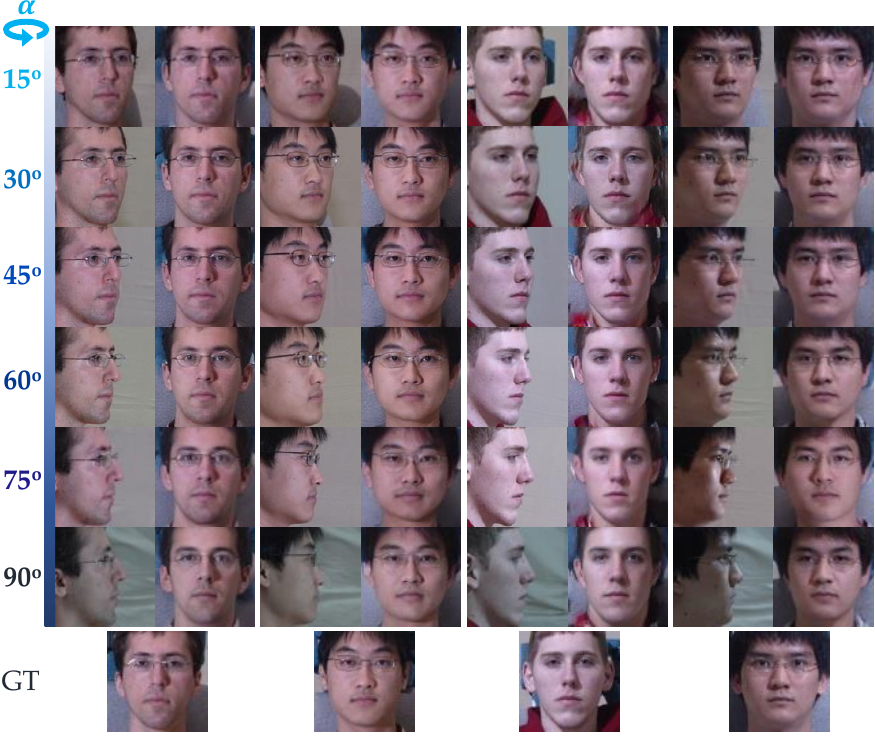} \\ 
    \vspace{-4mm}
    \caption{\textbf{Qualitative results.} DA-GAN synthesized results across a large range of yaw ($\alpha$) poses (\ie $15^\circ\sim90^\circ$).}
    \label{fig:facemontage}
    \vspace{-3mm}
\end{figure}

\subsubsection{Multi-scale Pixel-wise Loss}
Following \cite{li2019m2fpa}, we employ a multi-scale pixel-wise loss to constrain the content consistency. The multi-scale synthesized images are output by different layer of the decoder in \gls{g}.
The loss of the $i^{th}$ sample is the absolute mean difference of the multi-scaled synthesized and true frontal face (\ie $\hat{I}_i^{f}$ and $I_i^{f}$, respectively). Mathematically speaking:
 \begin{equation}
     \mathcal{L}_{pixel}  = \frac{1}{S}\sum_{s=1}^S \frac{1}{W_s H_s C} \sum_{w, h, c=1}^{W_s, H_s, C}\left| G(I_{s,w,h,c}^p)-I_{s,w,h,c}^f\right|,
 \end{equation}
where $S$ is the number of scales, $W_s$ and $H_s$ are the corresponding width and height of scale $s$. The synthesized frontal face $G(I_{s,w,h,c}^p)=\hat{I}_{s,w,h,c}^f$ is transformed by \gls{g} with learned parameters $\theta_G$. In our model, we set $S=3$, and the scales are $32\times32$, $64\times64$, and $128\times128$.

\subsubsection{Total Variation Regularization}
 A total variation regularization $\mathcal{L}_{tv}$ \cite{johnson2016perceptual} is also included to remove artifacts in synthesized images $\hat{I}^f$.
 \begin{equation}
    \mathcal{L}_{tv} = \sum_{c=1}^C \sum_{w,h=1}^{W,H} \left|\hat{I}_{w+1,h,c}^f - \hat{I}_{w,h,c}^f\right| + \left|\hat{I}_{w,h+1,c}^f - \hat{I}_{w,h,c}^f\right|,
\end{equation}
where $C, W, H$ denote the channel, width and height of $\hat{I}^f$.

\subsubsection{Overall Loss}
The objective function for the proposed is a weighted sum of aforementioned losses:
\begin{equation}
    \mathcal{L}_{G} = \lambda_1\mathcal{L}_{ID} + \lambda_2\mathcal{L}_{pixel} + \lambda_3\mathcal{L}_{adv} + \lambda_4\mathcal{L}_{tv},
\end{equation}
where $\lambda_1$, $\lambda_2$, $\lambda_3$, and $\lambda_4$ are hypter-parameters that control the trade-off of the loss terms. Detailed training algorithm of DA-GAN is provided in the supplementary material.

\section{Experiment}
We now demonstrate the proposed in photo-realistic face frontalization and pose invariant representation learning.

\begin{figure}[t]
    \centering
    \includegraphics[trim=0in 0.5in 0in 0in,clip,width=0.766\linewidth]{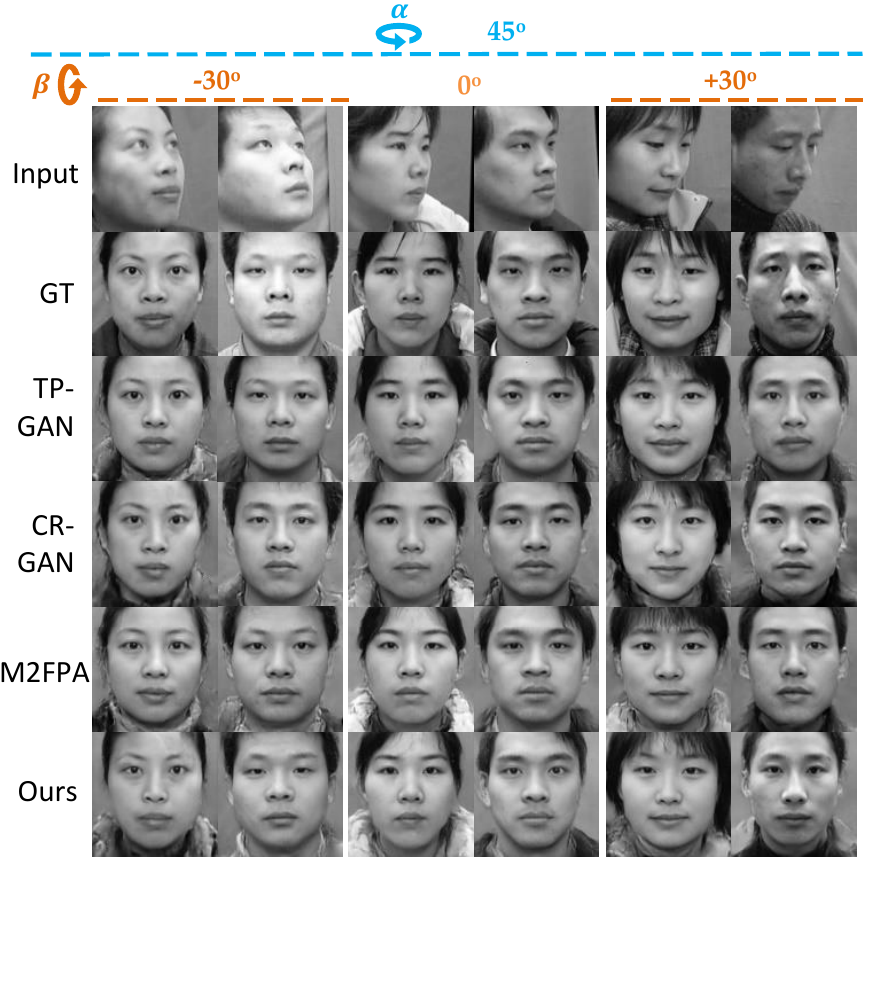}
    \caption{\textbf{CAS-PEAL-R1 results.} Comparison with SOTA on constant yaw ($\alpha$) and varying pitch ($\beta$) angles.}
    \label{fig:compare_state_of_the_art_cas}
    \vspace{-3mm}
\end{figure}

\subsection{Experiment Settings}
\subsubsection{Dataset}
\label{sec:Dataset}
The Multi-PIE dataset~\cite{gross2010multi} is the largest public database for face synthesis and recognition in the controlled setting. It consists of 337 subjects involved in up to 4 sessions.
We follow the second setting in \cite{li2019m2fpa, tian2018cr, yim2015rotating} to emphasize pose, illumination, and session (\ie time) variations. This setting includes images with neutral expressions from all four sessions and of the 337 identities. We use the images of the first 200 subjects for training, which includes samples with 13 poses within $\pm90^\circ$ and 20 illumination levels. Samples of the remaining 137 identities make-up the testing set, while samples neutral in expression and illumination make-up the gallery.
Note that there are no overlap subjects between training and test sets.

The CAS-PEAL-R1 dataset \cite{gao2007cas} is a public released large-scale Chinese face database with controlled pose, expression, accessory, and lighting variations. It contains 30,863 grayscale images of 1,040 subjects (595 males and 445 females). We only use images with various poses including 6 yaw angles (\ie $\alpha = \{0^\circ, \pm15^\circ, \pm30^\circ, \pm45^\circ\}$), 3 pitch angles (\ie $\beta = \{0^\circ, \pm30^\circ\}$), and a total of 21 yaw-pitch rotations. 
We use the first 600 subjects for training and the remaining 440 subjects for testing. 

LFW~\cite{huang2008labeled} contains 13,233 face images collected in unconstrained environment. It will be used to evaluate the frontalization performance in uncontrolled settings.

\subsubsection{Implementation Details}
To train our model, pairs of images $\{I^p, I^f\}$ consisting of one side-view image and corresponding frontal face image are required. We first cropped all images to a canonical view of size 128$\times$128 following \cite{li2019m2fpa}.
For MultiPIE, both real and generated images are RGB images. The identity preserving network used is pre-trained on MS-Celeb-1M \cite{guo2016ms} and fine-tuned on the training set of Multi-PIE.
For CAS-PEAL-R1, all images are set to grayscale. The identity preserving network used for training CAS-PEAL-R1 is pre-trained on grayscale images from MS-Celeb-1M. We set $\lambda_{1}=0.1$, $\lambda_{2}=10$, $\lambda_3=0.1$, $\lambda_4=1^{-4}$.

\begin{figure}[t]
    \centering
    \includegraphics[trim=0in 0in 0in 0in,clip,width=0.85\linewidth]{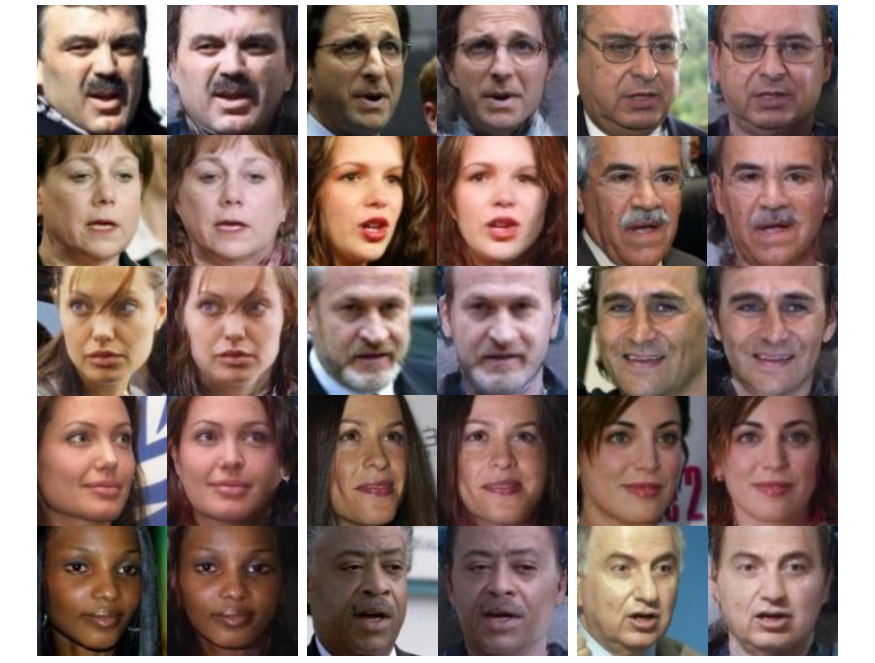}
    \caption{\textbf{Face Samples}. Results on LFW.} 
    \label{fig:lfw}
    \vspace{-1mm}
\end{figure}

\begin{table}[b]
    \centering  
    \caption{\textbf{LFW benchmark.} Face verification accuracy (ACC) and area-under-curve
    (AUC) results on LFW.}
    \begin{adjustbox}{max width=0.56\linewidth}
    \begin{tabular}{c|c|c}
    \hline
      & ACC (\%) & AUC (\%)\\
    \hline
    LFW-3D\cite{hassner2015effective} & 93.62   & 88.36 \\
    LFW-HPEN\cite{zhu2015high} & 96.25  & 99.39 \\
    FF-GAN\cite{yin2017towards} & 96.42   & 99.45 \\
    CAPG-GAN\cite{hu2018pose} & 99.37   & 99.90 \\
    M$^2$FPA\cite{li2019m2fpa} & 99.41   & 99.92 \\
    \hline
    Ours &  99.56  & 99.91\\
    \hline
    \end{tabular}
    \label{tab:lfw_face_recognition}
    \end{adjustbox}
    \vspace{-2mm}
\end{table}

\subsection{Face Synthesis}\label{sec:facesynthesis}
In this section, we visually compare the synthesized results of DA-GAN with state-of-the-art methods. 
Fig.~\ref{fig:compare_state_of_the_art} shows the qualitative comparison on MultiPIE. Specifically, we show the synthesis results of different methods under the pose of $60^\circ$ and $90^\circ$ to demonstrate the superior performance of the proposed DA-GAN on large poses.
Qualitative results show that the proposed DA-GAN recovers frontal images with finer detail (\ie more accurate facial shapes and textures), while the other methods tend to produce frontal faces with more inaccuracies. To show the realism of images synthesized from arbitrary views, Fig.~\ref{fig:facemontage} shows the synthesized frontal results of DA-GAN with various poses.

To further verify the improved results of DA-GAN across multiple yaws and pitches, we also compare results on the CAS-PEAL-R1 dataset, as it includes large pose variations. Since there is not much literature that have reported results on this data, we train and evaluate all the models on the same train and test splits of CAS-PEAL-R1 (Section~\ref{sec:Dataset}). To compare results, we used the public code of \gls{tpgan} and CR-GAN, and also implemented M$^2$FPA, as there was code available.
Fig.~\ref{fig:compare_state_of_the_art_cas} shows that our method generates the most realistic faces (\ie finer details), while preserving identity.

\begin{figure}[t]
    \centering
    \includegraphics[trim=0in 2.1in 0in 0in,clip,width=\linewidth]{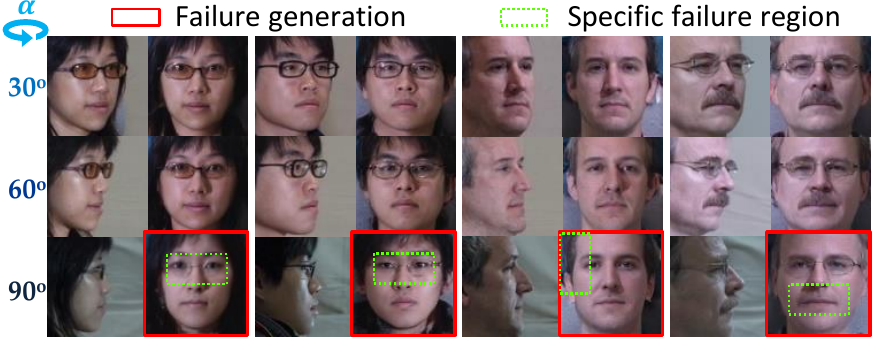}
    \caption{\textbf{Failure cases.} Instances of certain face attributes (\eg glasses, hair, and beard) fail to recover well from $90^\circ$. But those attributes can be recovered well from $30^\circ$ and $60^\circ$.}
    \label{fig:failure_cases}
    \vspace{-1mm}
\end{figure}

We show that DA-GAN can generate compelling results in most cases (Fig.~\ref{fig:samples}, \ref{fig:compare_state_of_the_art} and \ref{fig:facemontage}). But in some cases with extreme poses angle (\ie $90^\circ$) and large illumination discrepancy, sometimes it is difficult to recover frontal face images. We provide additional results in these challenging scenarios including some failure cases. 
As shown in Fig.~\ref{fig:failure_cases}, all face attributes can be well captured and recovered for poses of $30^\circ$ and $60^\circ$, while there are few cases that some of the face attributes (\eg eye-glasses, hair, and mustache) are not recovered well from a pose of $90^\circ$. Since those attributes are barely visible at $90^\circ$, the input side-view faces cannot provide enough information to synthesize correct frontal faces. In those cases, our model is incapable of synthesizing the exact frontal faces as the ground truth, but it can still generate reasonable and realistic results.

\begin{table}[b]
    \centering
    \glsunset{g}
    \glsunset{d}
    \caption{\textbf{MultiPIE benchmark.} Rank-1 recognition performance (\%) across views.}
    \begin{adjustbox}{max width=\linewidth}
    \begin{tabular}{c|ccccccc}
    \toprule
    & $\pm 90^\circ$ & $\pm 75^\circ$ & $\pm 60^\circ$ & $\pm 45^\circ$ & $\pm 30^\circ$ & $\pm 15^\circ$ & Avg\\
    \midrule
    TP-GAN~\cite{huang2017beyond}& 64.64 & 77.43 & 87.72 & 95.38 & 98.06 & 98.68 & 86.99\\ 
    FF-GAN~\cite{yin2017towards}& 61.20 & 77.20 & 85.20 & 89.70 & 92.50 & 94.60 & 83.40\\ 
    CAPGGAN~\cite{hu2018pose}& 66.05 &  83.05 & 90.63 & 97.33 & 99.56 & 99.82 & 89.41\\ 
    PIM1~\cite{zhao2018towards} & 71.60& 92.50 &97.00 &98.60 &99.30 &99.40&93.07\\
    PIM2~\cite{zhao2018towards} &75.00& 91.20& 97.70 &98.30& 99.40& 99.80&93.57\\
    M2FPA~\cite{li2019m2fpa}& 75.33& 88.74 & 96.18& \textbf{99.53}  & 99.78 & 99.96 &93.25\\ 
    \midrule
    Baseline&  66.08 &84.21& 90.84 & 97.71  & 99.25  &99.70&89.63\\ 
    Ours (\gls{g}+self-attention)&76.53 & 89.03 & 95.28 & 98.78 & 99.72 & \textbf{99.99} & 93.22\\ 
    Ours (\gls{d}+face-attention)&77.21 & 90.78 & 96.08 & 99.00 & 99.77 &  \textbf{99.99} & 93.81\\ 
    Ours (+dual-attention) &\textbf{81.56}  &  \textbf{93.24}& \textbf{97.27}&  99.15 & \textbf{99.88} & 99.98 & \textbf{95.18}\\
    \bottomrule
    \end{tabular}\label{tbl:rank1_multipie} 

    \end{adjustbox}
    \vspace{-2mm}
\end{table}

\begin{figure*}[t!]
 \centering
   \begin{subfigure}[b]{0.42\linewidth}
    \includegraphics[trim=0.1in 2.23in 0.1in 0in,clip,width=\linewidth]{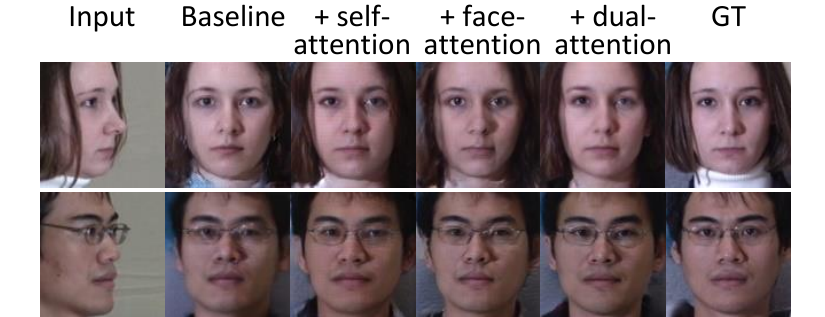}
    \caption{Attention-level}
    \label{fig:ablation_attention}
  \end{subfigure}
  \begin{subfigure}[b]{0.42\linewidth}
    \includegraphics[trim=0.1in 0.74in 0.1in 1.44in,clip,width=\linewidth]{ablation4.pdf}
    \caption{Mask-level}
    \label{fig:ablation_mask}
  \end{subfigure}
\caption{\textbf{Ablation Study (qualitative results).} Frontalization results generated by variation models with removed components in (a) attention-level and (b) mask-level.}
\label{fig:ablationStudy}
\end{figure*}
\begin{table*}[t]
    \centering
    \glsunset{g}
    \glsunset{d}
    \caption{\textbf{CAS\_PEAL\_R1 benchmark.} Rank-1 recognition performance (\%).}
    \begin{adjustbox}{max width=0.85\linewidth}
        \begin{tabular}{c|ccccc|cccc|ccccc}
        \toprule
        & \multicolumn{5}{c|}{Pitch ($-15^\circ$)} & \multicolumn{4}{c|}{Pitch ($0^\circ$)} & \multicolumn{5}{c}{Pitch ($+15^\circ$)} \\
        Yaw& $\pm 0^\circ$ & $\pm 15^\circ$ & $\pm 30^\circ$ & $\pm 45^\circ$ & Avg\_1 & $\pm 15^\circ$ & $\pm 30^\circ$ & $\pm 45^\circ$ & Avg\_2 & $\pm 0^\circ$ & $\pm 15^\circ$ & $\pm 30^\circ$ & $\pm 45^\circ$ & Avg\_3 \\
        \midrule
    
        TP-GAN~\cite{huang2017beyond}& 98.86 & 98.94 & 98.89 & 97.62 & 98.58 & \textbf{100.00} & 99.94 & 98.71 & 99.55 & 97.68& 97.73& 97.45 & 95.83 & 97.17\\ 
        CR-GAN~\cite{zhao2018towards}& 83.98 & 83.91 & 83.17 & 80.38 & 82.86 & 97.61 & 95.80 & 89.73 & 94.38 & 89.74 & 89.44 & 87.95 & 83.90 & 87.76\\ 
        M2FPA~\cite{li2019m2fpa}& 99.38 & 99.42 & 99.30 & 98.53 & 99.16 & \textbf{100.00} & 99.94
        & 99.36 & 99.77 & 98.60 & 98.69 & 98.58 & 97.84 & 98.43\\ 
        DA-GAN &\textbf{99.71}  & \textbf{99.72}& \textbf{99.65}&  \textbf{98.99}  & \textbf{99.52} & \textbf{100.00} &\textbf{100.00} &\textbf{99.70} & \textbf{99.90} &\textbf{98.96}&\textbf{98.98}&\textbf{98.86}&\textbf{98.13} & \textbf{98.73}\\
        \bottomrule
        \end{tabular}\label{tbl:rank1_cas} 
    \end{adjustbox}
    \vspace{-2mm}
\end{table*}

\begin{table}[b]
    \centering
    \glsunset{g}
    \glsunset{d}
    \caption{\textbf{Ablation study: quantitative results.} Rank-1 recognition performance (\%) across views.}\label{tab:ablation}
    \begin{adjustbox}{max width=0.88\linewidth}
    \begin{tabular}{c|ccccccc}
    \toprule
    & $\pm 90^\circ$ & $\pm 75^\circ$ & $\pm 60^\circ$ & $\pm 45^\circ$ & $\pm 30^\circ$ & $\pm 15^\circ$ & Avg\\
    \midrule
    \gls{d}+$D_{h}$& 72.23 & 85.58 & 92.96 & 98.38 & \textbf{99.81} & \textbf{99.99} & 91.49\\ 
    \gls{d}+$D_{s}$& 72.25  &  87.68  &  94.0 & 98.62 & 99.69 & 99.98 & 92.04\\ 
    \gls{d}+$D_{k}$& \textbf{77.23} & 88.33 & 95.01  & 98.83 & 99.72 & \textbf{99.99} & 93.19\\
    \midrule
    \gls{d}+face-attention&77.21  &  \textbf{90.78}  &  \textbf{96.08} & \textbf{99.00} & 99.77 & \textbf{99.99} & \textbf{93.81} \\ 
    \bottomrule
\end{tabular}\label{tbl:rank1_ablation} 
    \end{adjustbox}
    \vspace{-3mm}
\end{table}

\subsection{Identity Preserving Property}\label{sec:identity}
To quantitatively demonstrate the identity preserving ability of proposed DA-GAN, we evaluate face recognition accuracy on synthesized frontal images. Table~\ref{tbl:rank1_multipie} compares performance with existing state-of-the-art on MultiPIE across different poses. Results are reported with rank-1 identification rate. We employ a pre-trained 29-layer Light-CNN\cite{wu2018light} as the face recognition model to extract features and use cosine-distance metric to compute the similarity of these features.
Larger pose tends to provide less information, making preserving the identity in the synthesized difficult. As shown in Table~\ref{tbl:rank1_multipie}, the performance of existing methods sharply drops as pose degree increases to $75^\circ$ and larger, while our method still have compelling performance at these extreme poses (\ie $75^\circ$ and $90^\circ$). Besides, DA-GAN can also achieves the best or comparable performace across other smaller poses (\ie $15^\circ$ $\sim$ $60^\circ$).
Similarly, Table~\ref{tbl:rank1_cas} shows the rank-1 identification rate for CAS-PEAL-R1 across yaw ($\alpha$) and pitch ($\beta$) pose variations. The results are summarized in Table~\ref{tbl:rank1_cas}, which consistently demonstrates the superior identity preserving ability of DA-GAN across multiple poses. 

We analyze, quantitatively, the benefits of using the proposed in the LFW benchmark (Table~\ref{tab:lfw_face_recognition}).
Specifically, face recognition performance is evaluated on synthesized frontal images. The results of the state-of-the-art methods in Table~\ref{tab:lfw_face_recognition} are from \cite{li2019m2fpa}. The qualitative results of LFW are in Fig.~\ref{fig:lfw}.

\subsection{Ablation Study}\label{sec:ablation}
The contributions of self-attention in \gls{g} and face-attention in \gls{d} to the frontalized performance are analyzed via ablation studies. Our baseline model only consists of a U-Net generator~\cite{ronneberger2015u} and one ordinary frontal face discriminator. In other words, the baseline model is the proposed DA-GAN without any attention schemes. The other two variants are constructed by adding the self-attention or face-attention solely to the baseline model, while the proposed DA-GAN has dual attentions. Besides the ablation study on the face attention mechanism as one, we also characterize each discriminator used in the face attention scheme. 

\subsubsection{Effects of two types of attentions}
To highlight the importance of self-attention in \gls{g} and face-attention in \gls{d}, Table~\ref{tbl:rank1_multipie} shows the quantitative comparison between the proposed method and its variants with different attentions.
Results show that using either of the attentions will significantly boost the performance of recognition, while employing both of them together will achieve the best performance, especially for large poses.
The face recognition results signified that DA-GAN improves the recognition accuracy of extreme poses (\ie $90^\circ$) up to 23.43\% when compared to the its variants using a subset of the attention types.

Furthermore, we show qualitative comparisons between the proposed method and its variants of incomplete attentions (see Fig.~\ref{fig:ablation_attention}). The synthesized results of adding self-attention have relatively less blurriness (\eg fuzzy face and ear) than those with face-attention and with dual-attention. However, from visualization aspect, it has comparable identity preserving ability with dual-attention model. 
In contrast, the model with face-attention alone can produce photo-realistic faces, but preserves less identity information.
By introducing the two different types of attention in generator and discriminator separately, our DA-GAN can generate identity preserving inference of frontal views with relatively more details (\ie facial appearance and textures).

\subsubsection{Effects of different masks employed in \gls{d}}
We also explore the contributions of three different masks used as face attention in the \gls{d} quantitatively (Table~\ref{tbl:rank1_ablation}) and qualitatively (Fig.~\ref{fig:ablation_mask}). 
Quantitative results show that key-point features contribute the most to face recognition task, and the hairline features contribute the least. 
Furthermore, we show qualitative comparisons between variants of different masks. By adding hair discriminator ($D_h$), the model can generate relatively sharper edges in hair region. Skin discriminator ($D_s$) helps with low-frenquecy features, and key-point discriminator ($D_k$) helps generating faithful facial attributes (\eg eyes) to ground-truth. Finally, we gain complementary information by combing all of them as face-attention.

\section{Conclusions}
We achieved state-of-the-art with a novel frontal face synthesizer: namely, DA-GAN, which introduced self-attention in the \gls{g} that was then trained in an adversarial manner via a \gls{d} equipped with face-attention. During inference, the proposed framework effectively synthesized faces from up to 90$^\circ$ faces to a frontal view. Furthermore, the visually appealing results carry practical significance (\ie face recognition systems typically improve with improved alignment done during the preprocessing stage). We perceptually and numerically demonstrated that our method synthesized compelling results, and improved the facial recognition performance.

\bibliographystyle{ieee}
\bibliography{ieee}

\end{document}